\pdfoutput=1

\documentclass{book}
\usepackage{pdfpages}
\usepackage{graphicx}
\usepackage{fancyhdr}
\usepackage{hyperref}
\usepackage[utf8]{inputenc}
\usepackage[T1]{fontenc}

\usepackage{amssymb, amstext, amsmath, epstopdf, booktabs, verbatim, gensymb, geometry, natbib, lmodern}
\usepackage{tocloft}
\newlength{\standardchapnumwidth}
\AtBeginDocument{\setlength{\standardchapnumwidth}{\cftchapnumwidth}}

\newcommand*\cpiType{Volume 1}
\newcommand*\Date{June 2020}
\newcommand*\Author{Federico Castagna\\ Francesca Mosca\\ Jack Mumford\\ Ștefan Sarkadi \\Andreas Xydis}

\definecolor{myblue}{HTML}{154360}
\definecolor{emerald}{HTML}{3cb371}

\begin{document}

\newgeometry{margin = 0in}


\pagecolor{emerald}
\setlength{\fboxsep}{0pt}
\hfill \colorbox{myblue}{\makebox[3.22in][r]{\shortstack[r]{\vspace{3.3in}}}}%
\setlength{\fboxsep}{15pt}
\setlength{\fboxrule}{5pt}
\colorbox{white}{\makebox[\linewidth][c]{\includegraphics[width=1.3in]{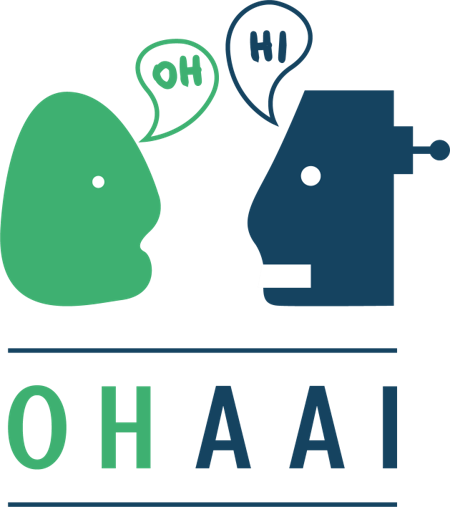}\hspace{0.35in} \shortstack[l]{\vspace{10pt}\fontsize{40}{40}\rmfamily\color{myblue} Online Handbook of \\
\vspace{10pt}
\fontsize{40}{40}\rmfamily\color{myblue} Argumentation for AI\\%
\fontsize{20}{20}\rmfamily\color{myblue} \cpiType}}}%
\setlength{\fboxsep}{0pt}
\vspace{-0.25pt}
\hfill \colorbox{myblue}{\hspace{.25in} \parbox{2.97in}{\vspace{2in} \color{white} \large{Edited by \\ \\ \Author  \\ \\  \Date \vspace{2.15in} \vfill}}}%
\restoregeometry

\nopagecolor

\thispagestyle{empty}
\pagenumbering{gobble}

\begin{center}
    \textbf{{\huge Preface}}
\end{center}

\hfill

This volume contains revised versions of the papers selected for the first volume of the Online Handbook of Argumentation for AI (OHAAI). Previously, formal theories of argument and argument interaction have been proposed and studied, and this has led to the more recent study of computational models of argument. Argumentation, as a field within artificial intelligence (AI), is highly relevant for researchers interested in symbolic representations of knowledge and defeasible reasoning. 
The purpose of this handbook is to provide an open access and curated anthology for the argumentation research community. OHAAI is designed to serve as a research hub to keep track of the latest and upcoming PhD-driven research on the theory and application of argumentation in all areas related to AI. The handbook’s goals are to:

\begin{enumerate}
    \item Encourage collaboration and knowledge discovery between members of the argumentation community.
    \item Provide a platform for PhD students to have their work published in a citable peer-reviewed venue.
    \item Present an indication of the scope and quality of PhD research involving argumentation for AI.
\end{enumerate}

The papers in this volume are those selected for inclusion in OHAAI Vol.1  following a back-and-forth peer-review process undertaken by the editors of OHAAI Vol.1. The volume thus presents a strong representation of the current state of the art research of argumentation in AI that has been strictly undertaken during PhD studies. Papers in this volume are listed alphabetically by author. We hope that you will enjoy reading this handbook.
\begin{flushright}
\noindent\begin{tabular}{r}
\makebox[1.3in]{}\\
\textit{Editors}\\
Federico Castagna\\
Francesca Mosca\\
Jack Mumford\\
Ștefan Sarkadi\\
Andreas Xydis\\\\
\textbf{June 2020}
\end{tabular}
\end{flushright}


\pagenumbering{gobble}

\begin{center}
    \textbf{{\huge Acknowledgements}}
\end{center}

\hfill

\noindent
We thank the senior researchers in the area of Argumentation and Artificial Intelligence for their efforts in spreading the word about the OHAAI project with early-career researchers.

\hfill

\noindent
We are also grateful to ArXiv for their willingness to publish this handbook.

\hfill

\noindent
We are especially thankful to Costanza Hardouin for designing the OHAAI logo.

\hfill

\noindent
We owe many thanks to Sanjay Modgil for helping to form the motivation for the handbook, and to Elizabeth Black and Simon Parsons for their advice and guidance that enabled the OHAAI project to come to fruition.

\hfill

\noindent
We owe special thanks to the contributing authors: Federico Castagna, Timotheus Kampik, Atefeh Keshavarzi Zafarghandi, Mickaël Lafages, Jack Mumford, Christos T. Rodosthenous, Samy Sá, Ștefan Sarkadi, Joseph Singleton, Kenneth Skiba, Andreas Xydis. Thank you for making the world of argumentation greater! 

\newgeometry{margin = 0.9in}

\pagenumbering{arabic}

\tableofcontents 
\thispagestyle{empty}
\clearpage




\pagestyle{fancy}
\addtocontents{toc}{\protect\renewcommand{\protect\cftchapleader}
     {\protect\cftdotfill{\cftsecdotsep}}}
\addtocontents{toc}{\setlength{\protect\cftchapnumwidth}{0pt}}

\refstepcounter{chapter}\label{1}
\addcontentsline{toc}{chapter}{Argument games for dialectical classical logic argumentation \\ \textnormal{\textit{Federico Castagna}}}
\includepdf[pages=-,pagecommand={\thispagestyle{plain}}]{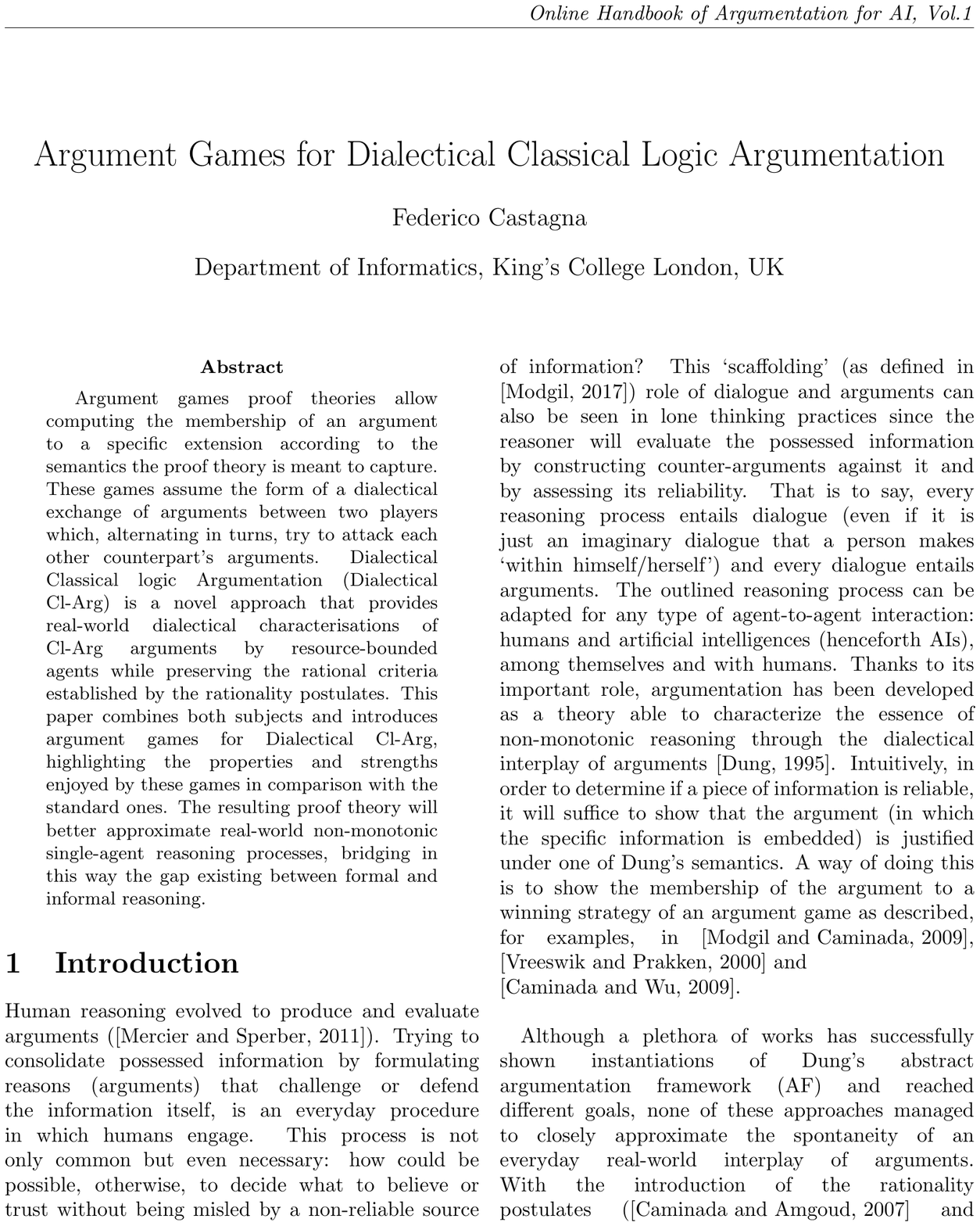}

\refstepcounter{chapter}\label{2}
\addcontentsline{toc}{chapter}{Economic rationality and abstract argumentation \\
\textnormal{\textit{Timotheus Kampik}}}
\includepdf[pages=-,pagecommand={\thispagestyle{plain}}]{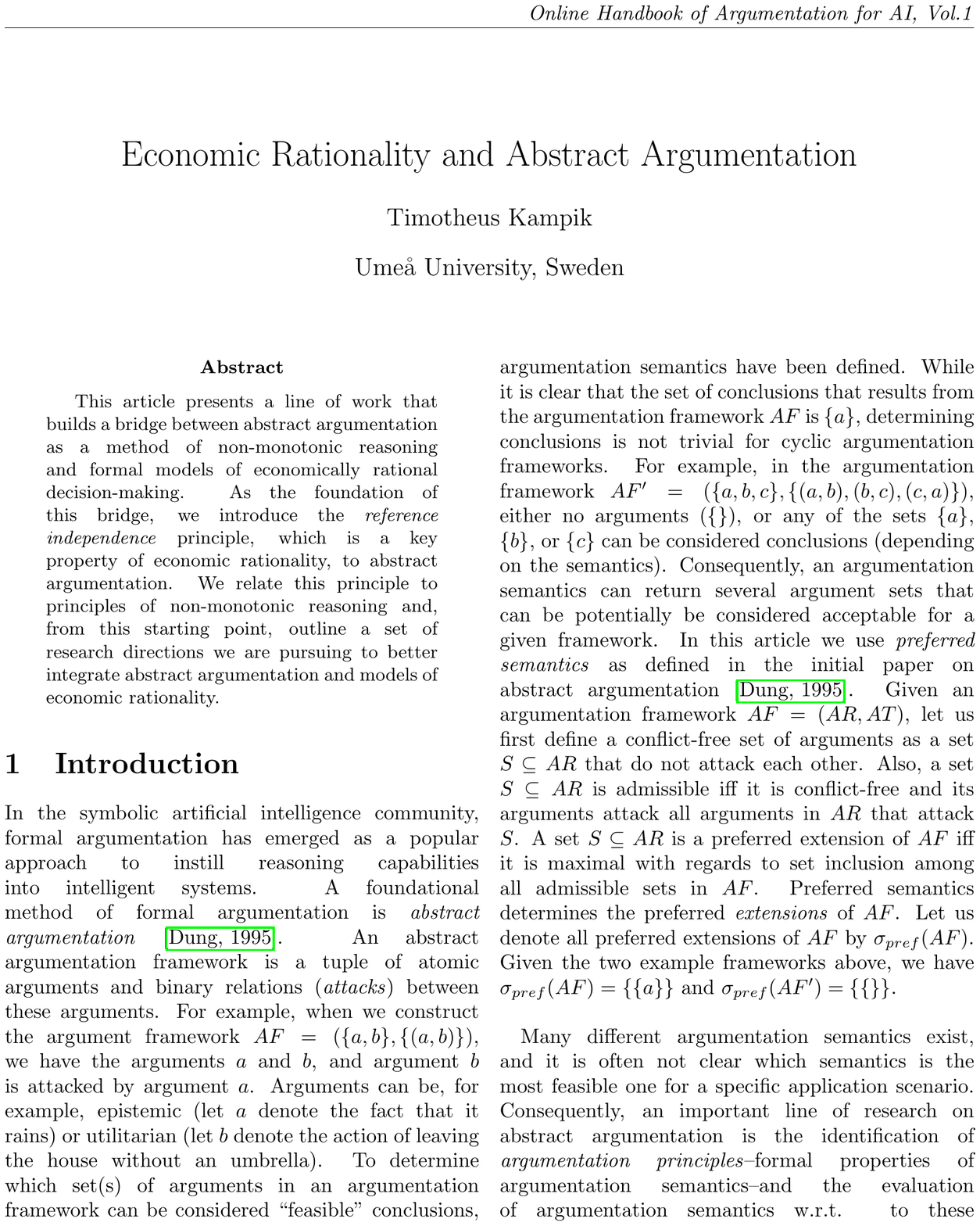}

\refstepcounter{chapter}\label{3}
\addcontentsline{toc}{chapter}{A discussion game for the credulous decision problem of abstract dialectical frameworks under preferred semantics \\
\textnormal{\textit{Atefeh Keshavarzi Zafarghandi}}}
\includepdf[pages=-,pagecommand={\thispagestyle{plain}}]{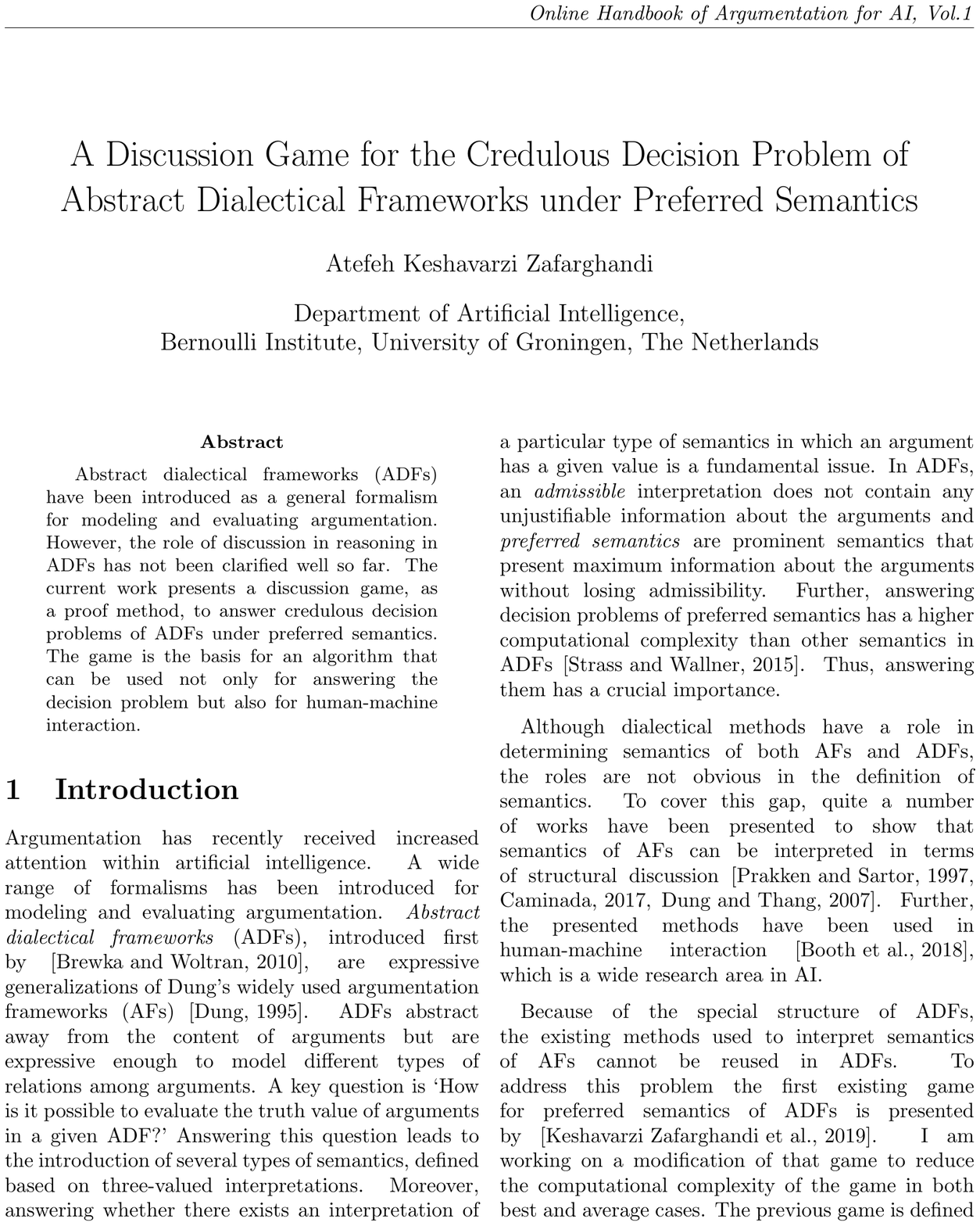}

\refstepcounter{chapter}\label{4}
\addcontentsline{toc}{chapter}{Algorithms and tools for abstract argumentation \\
\textnormal{\textit{Mickaël Lafages}}}
\includepdf[pages=-,pagecommand={\thispagestyle{plain}}]{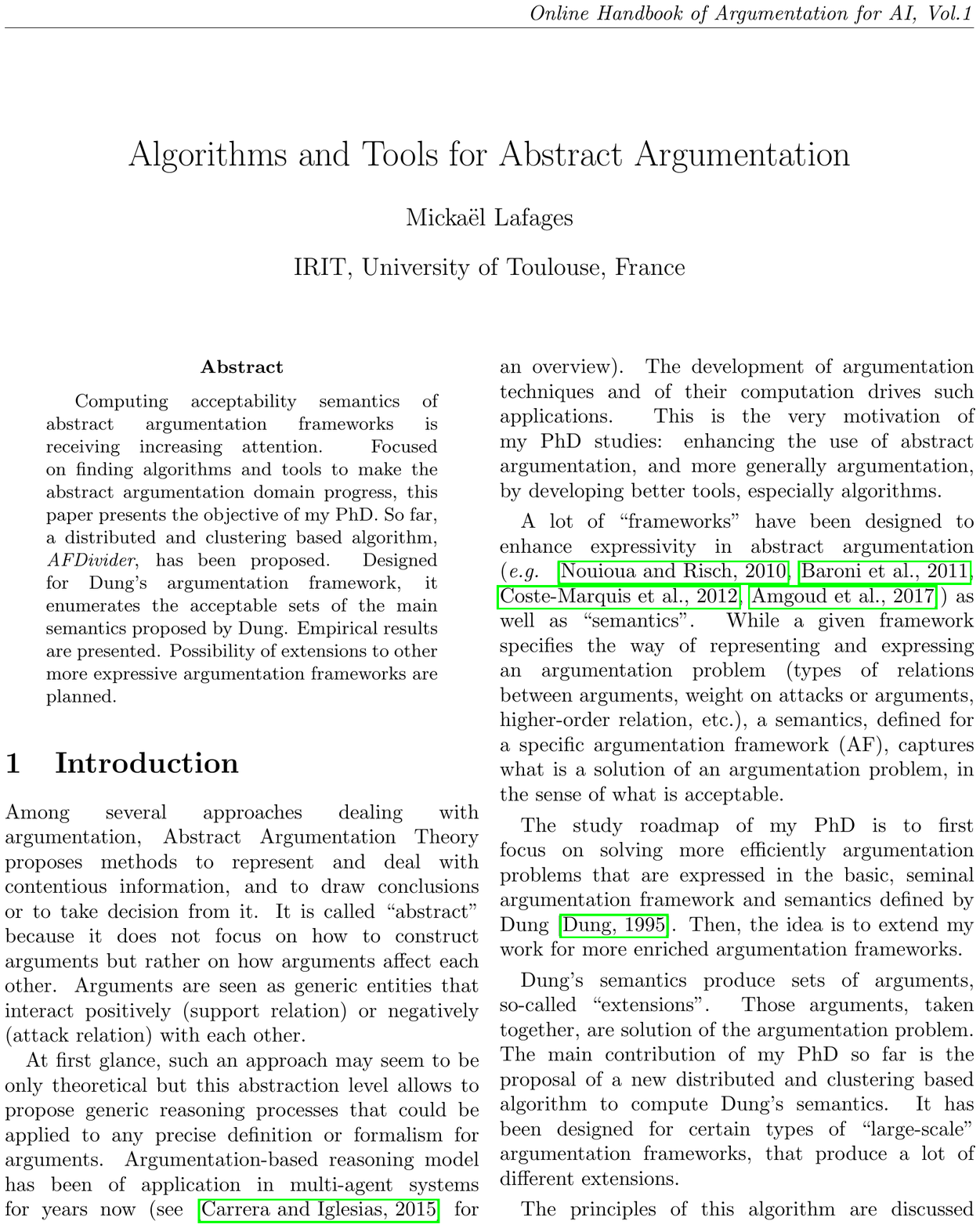}

\refstepcounter{chapter}\label{5}
\addcontentsline{toc}{chapter}{Crafting neural argumentation networks \\
\textnormal{\textit{Jack Mumford}}}
\includepdf[pages=-,pagecommand={\thispagestyle{plain}}]{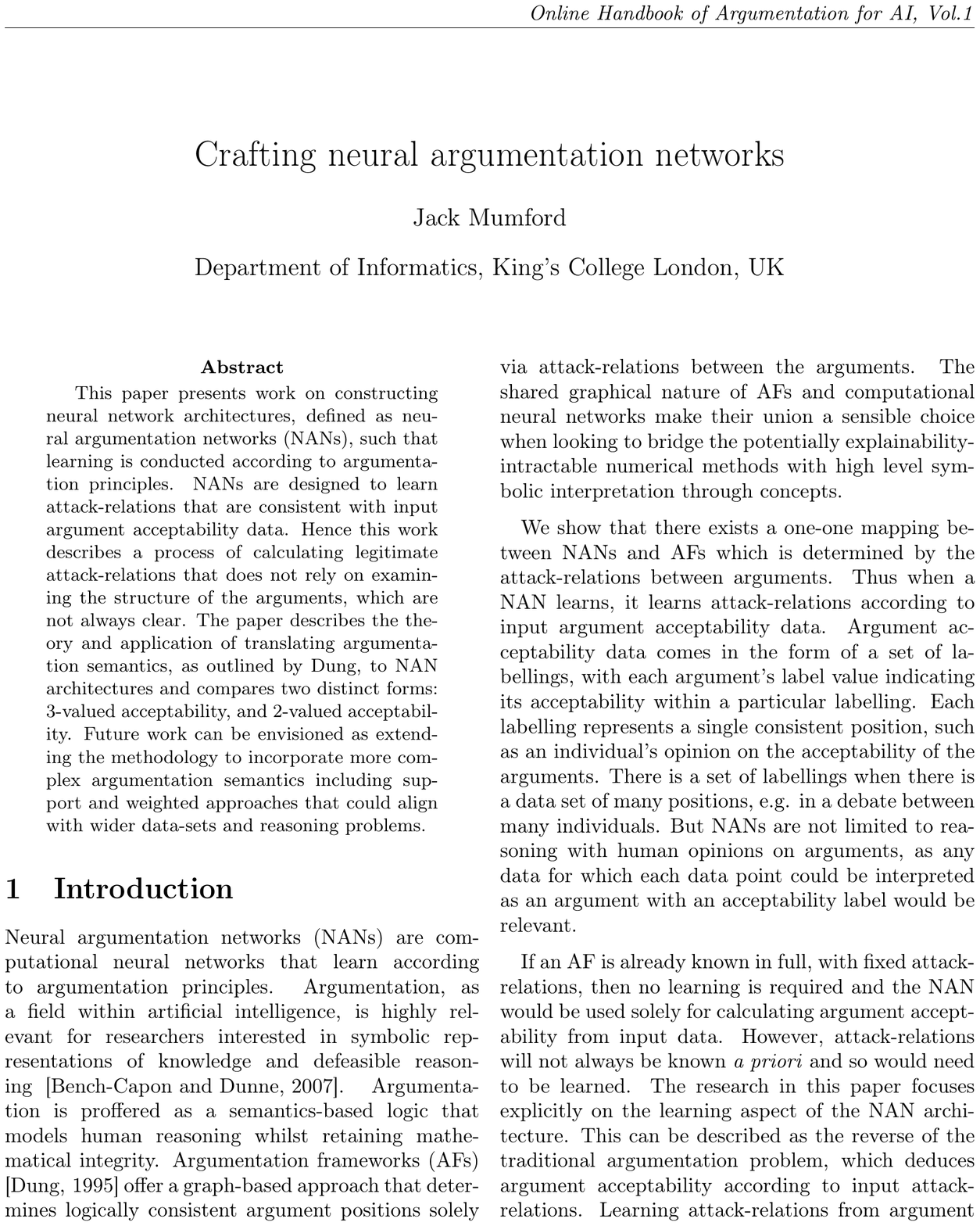}

\refstepcounter{chapter}\label{6}
\addcontentsline{toc}{chapter}{Understanding stories using crowdsourced commonsense knowledge \\
\textnormal{\textit{Christos T. Rodosthenous}}}
\includepdf[pages=-,pagecommand={\thispagestyle{plain}}]{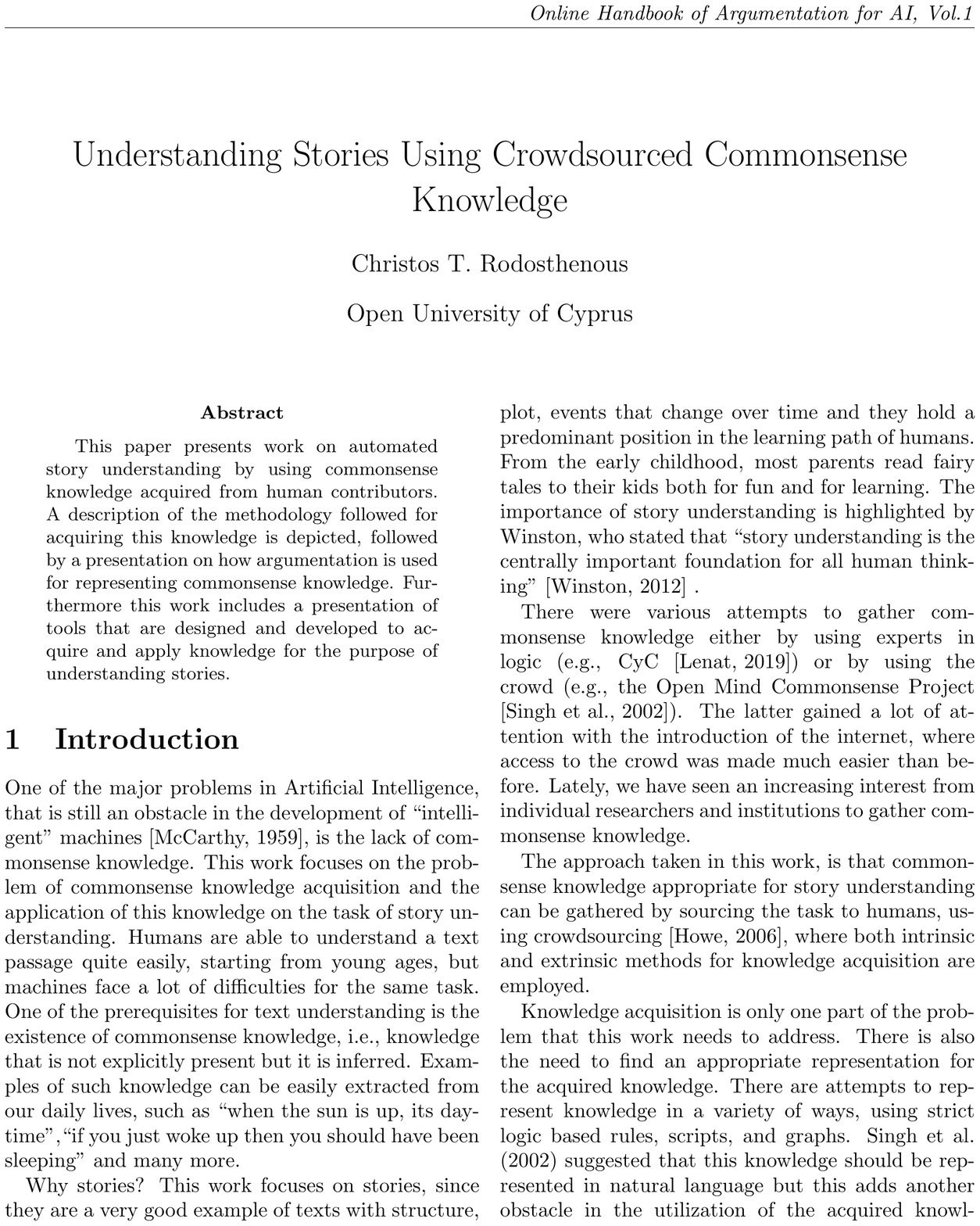}

\refstepcounter{chapter}\label{7}
\addcontentsline{toc}{chapter}{On the expressive power of argumentation formalisms \\
\textnormal{\textit{Samy Sá}}}
\includepdf[pages=-,pagecommand={\thispagestyle{plain}}]{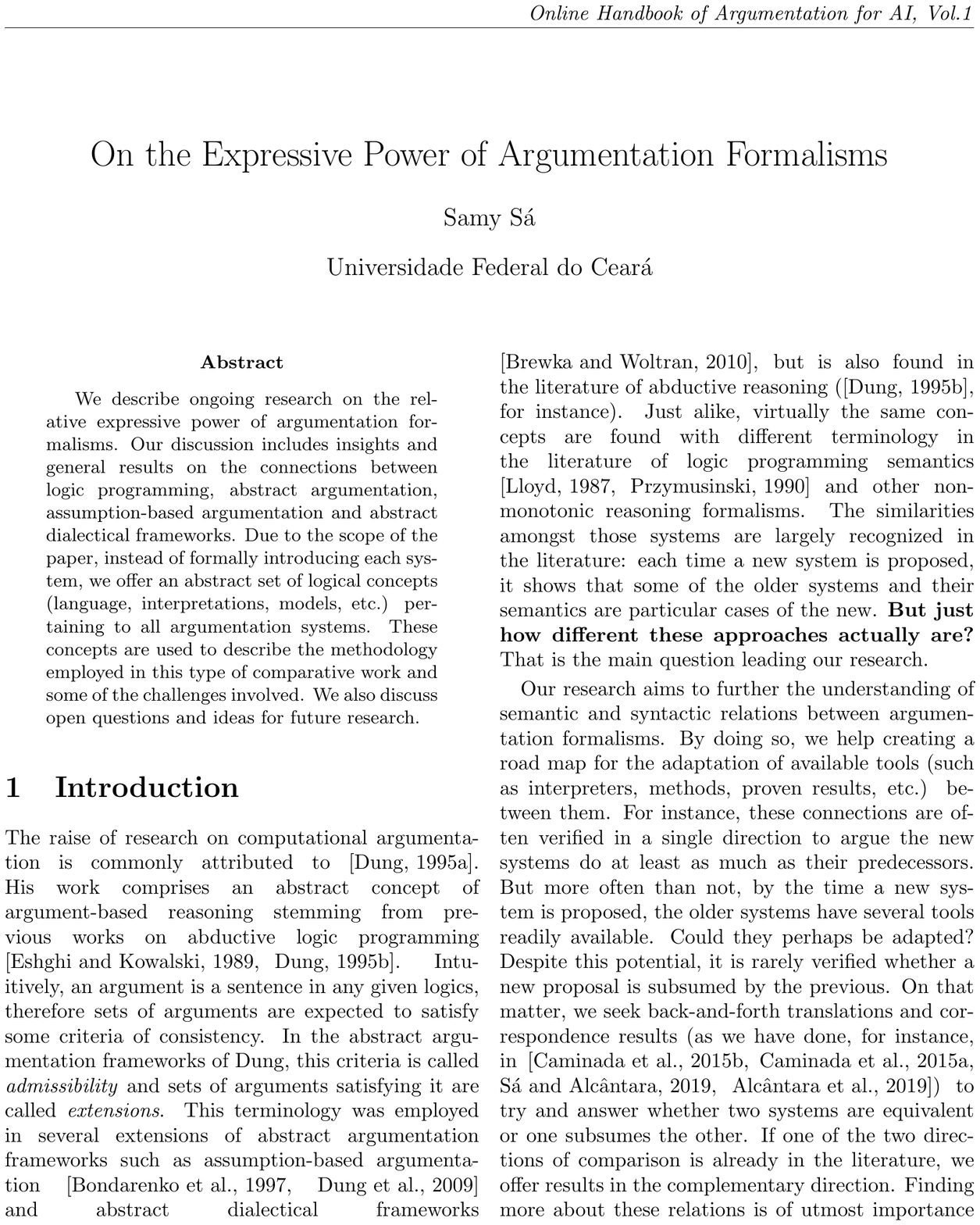}

\refstepcounter{chapter}\label{8}
\addcontentsline{toc}{chapter}{Argumentation-based dialogue games for modelling deception \\
\textnormal{\textit{Ștefan Sarkadi}}}
\includepdf[pages=-,pagecommand={\thispagestyle{plain}}]{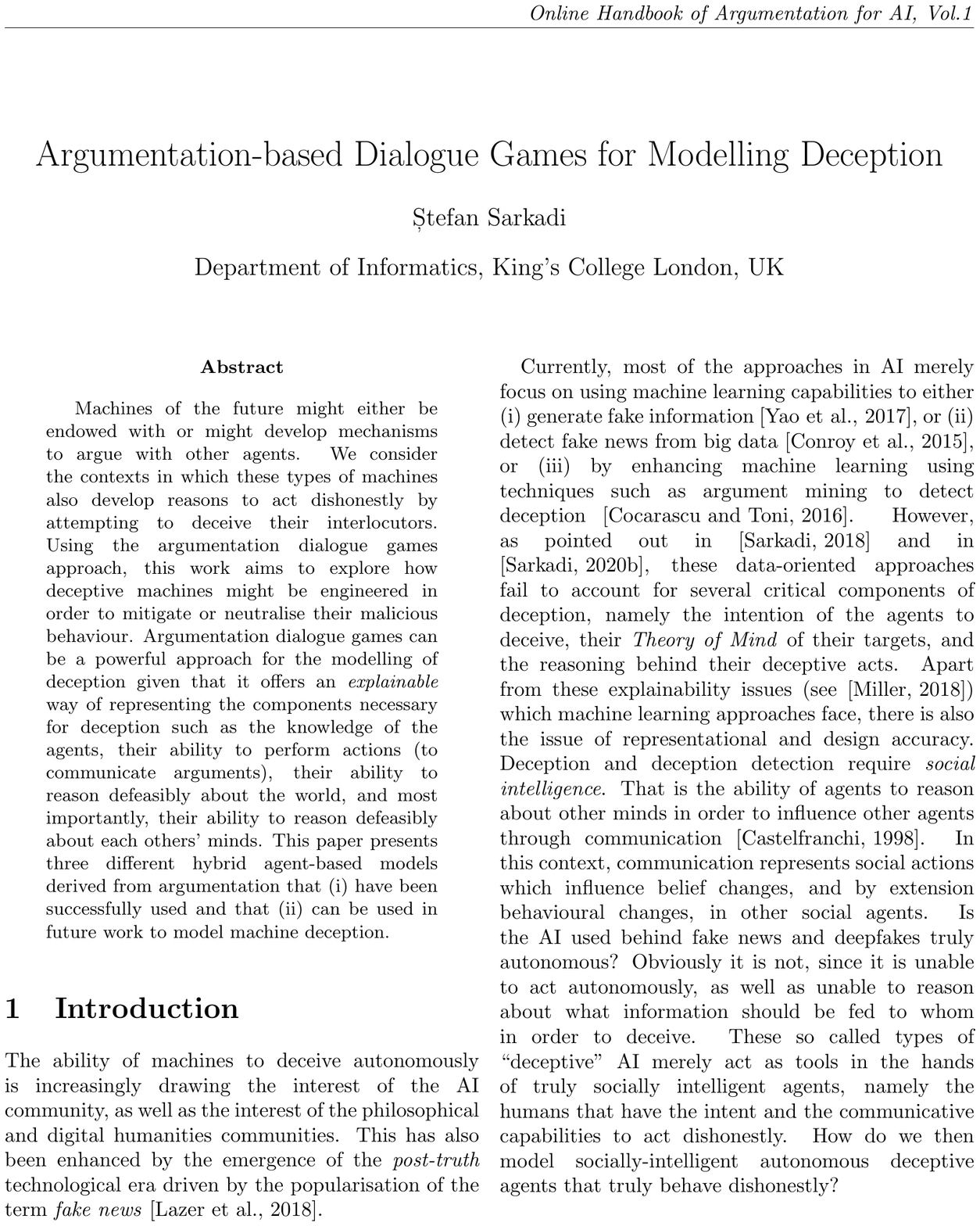}

\refstepcounter{chapter}\label{9}
\addcontentsline{toc}{chapter}{On the link between truth discovery and bipolar abstract argumentation \\
\textnormal{\textit{Joseph Singleton}}}
\includepdf[pages=-,pagecommand={\thispagestyle{plain}}]{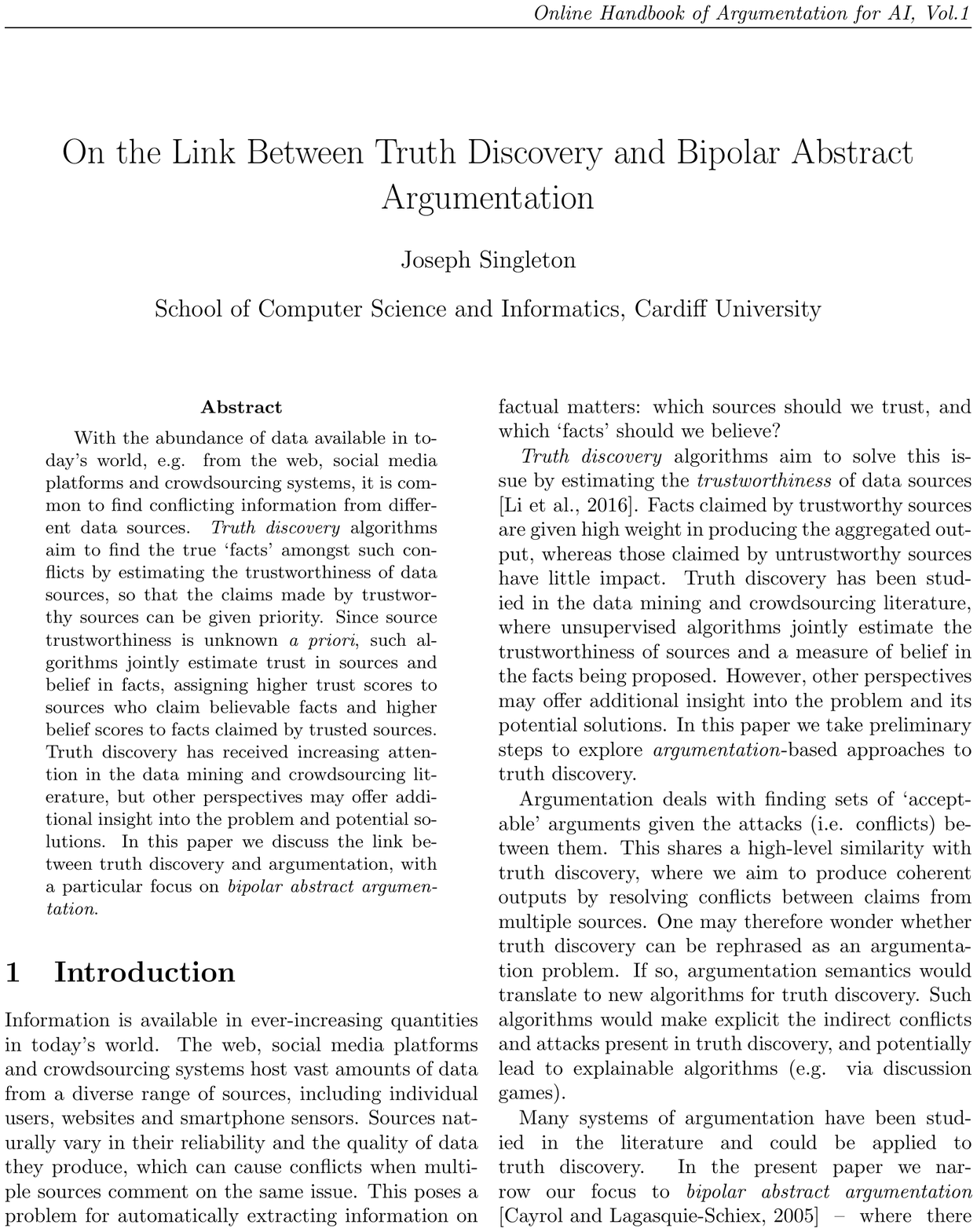}

\refstepcounter{chapter}\label{10}
\addcontentsline{toc}{chapter}{A first idea for a ranking-based semantics using system Z \\
\textnormal{\textit{Kenneth Skiba}}}
\includepdf[pages=-,pagecommand={\thispagestyle{plain}}]{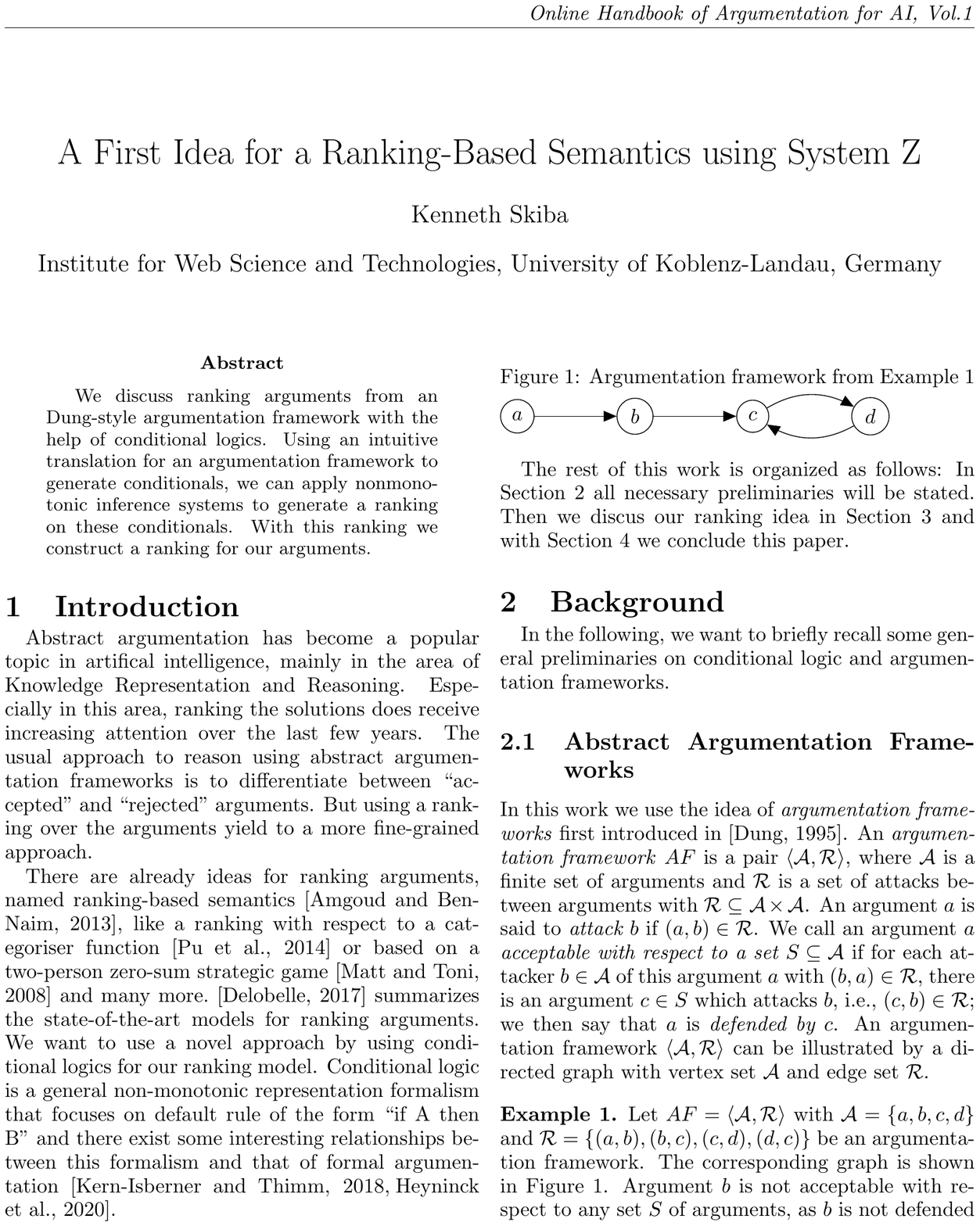}

\refstepcounter{chapter}\label{11}
\addcontentsline{toc}{chapter}{Speech acts and enthymemes in argumentation-based dialogues \\
\textnormal{\textit{Andreas Xydis}}}
\includepdf[pages=-,pagecommand={\thispagestyle{plain}}]{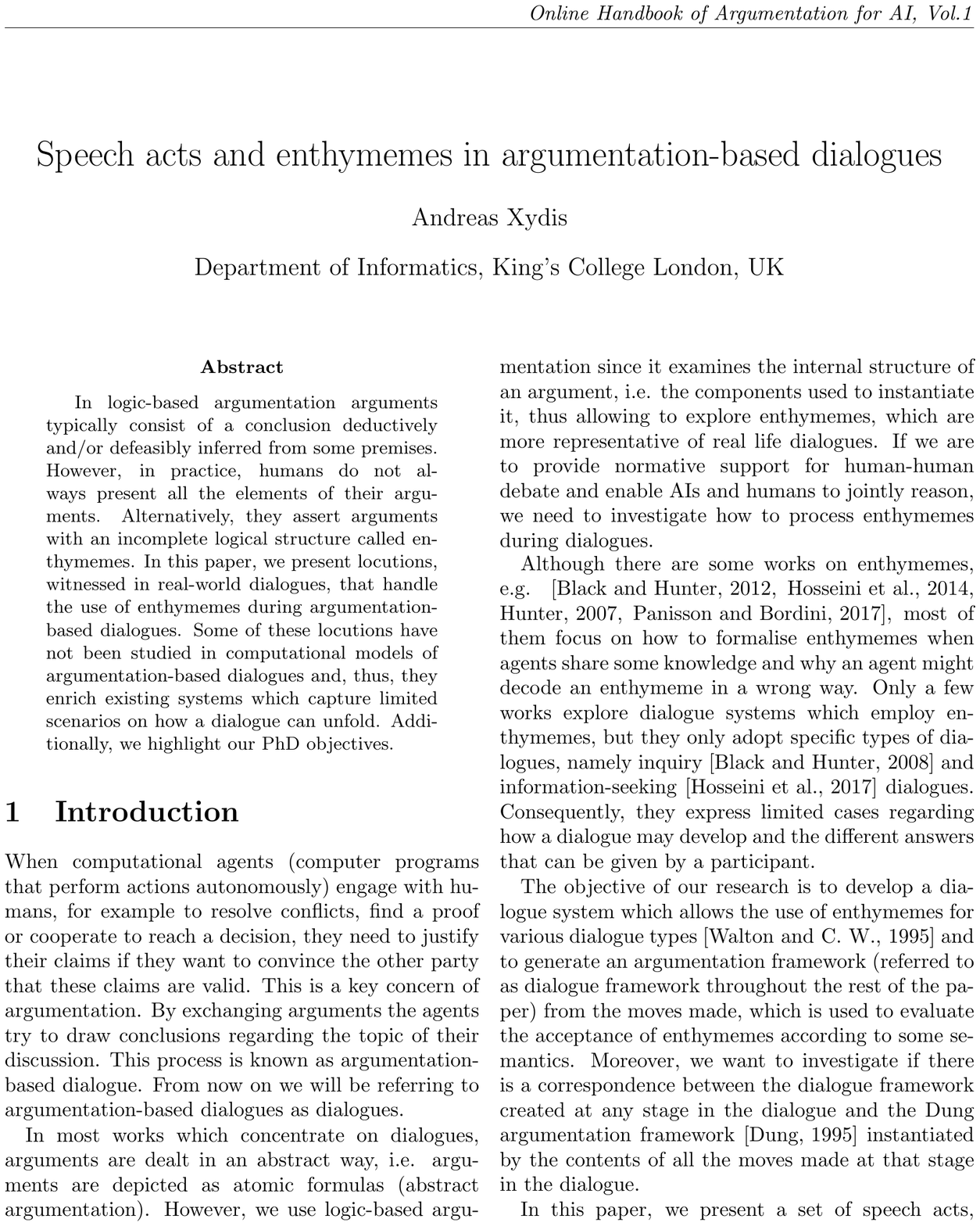}

\end{document}